\documentclass[nohyperref]{article}

\usepackage{microtype}
\usepackage{graphicx}
\usepackage{subfigure}
\usepackage{booktabs} % for professional tables

\usepackage{hyperref}

\usepackage[accepted]{icml2022}

\usepackage{amsmath}
\usepackage{amssymb}
\usepackage{mathtools}
\usepackage{amsthm}
\usepackage{bbm}
\usepackage{custom_commands}
\usepackage{bm}
\usepackage{enumitem}
\usepackage[noorphans,vskip=0ex]{quoting}
\usepackage{xcolor}
\usepackage{xspace}

\usepackage{etoolbox}
\newcommand{\zerodisplayskips}{%
  \setlength{\abovedisplayskip}{4pt}%
  \setlength{\belowdisplayskip}{4pt}%
  \setlength{\abovedisplayshortskip}{1pt}%
  \setlength{\belowdisplayshortskip}{1pt}}
\appto{\normalsize}{\zerodisplayskips}
\appto{\small}{\zerodisplayskips}
\appto{\footnotesize}{\zerodisplayskips}

\usepackage{titlesec}
\titlespacing\section{-5pt}{0pt plus 1pt minus 1pt}{0pt plus 1pt minus 1pt}
\titlespacing\subsection{-5pt}{0pt plus 1pt minus 2pt}{0pt plus 1pt minus 1pt}
\titlespacing\subsubsection{-5pt}{0pt plus 1pt minus 2pt}{0pt plus 1pt minus 1pt}

\usepackage[capitalize,noabbrev]{cleveref}

\theoremstyle{plain}
\usepackage[textsize=tiny]{todonotes}

\icmltitlerunning{Invariant-Feature Subspace Recovery}

\begin{document}

\twocolumn[
\icmltitle{Provable Domain Generalization via Invariant-Feature Subspace Recovery}



\icmlsetsymbol{equal}{*}

\begin{icmlauthorlist}
\icmlauthor{Haoxiang Wang}{uiuc}
\icmlauthor{Haozhe Si}{uiuc}
\icmlauthor{Bo Li}{uiuc}
\icmlauthor{Han Zhao}{uiuc}
\icmlaffiliation{uiuc}{University of Illinois at Urbana-Champaign, Urbana, IL, USA}

\end{icmlauthorlist}

\icmlcorrespondingauthor{Haoxiang Wang}{hwang264@illinois.edu}

\icmlkeywords{Machine Learning, ICML}

\vskip 0.3in
% \vspace{-3em}
]



\printAffiliationsAndNotice{}  % leave blank if no need to mention equal contribution

\begin{abstract}
Domain generalization asks for models trained over a set of training environments to perform well in unseen test environments. Recently, a series of algorithms such as Invariant Risk Minimization (IRM) has been proposed for domain generalization. However, \citet{risks-of-IRM} shows that in a simple linear data model, even if non-convexity issues are ignored, IRM and its extensions cannot generalize to unseen environments with less than $d_s\mathrm{+}1$ training environments, where $d_s$ is the dimension of the spurious-feature subspace. In this paper, we propose to achieve domain generalization with \textbf{I}nvariant-feature \textbf{S}ubspace \textbf{R}ecovery (ISR). Our first algorithm, ISR-Mean, can identify the subspace spanned by invariant features from the first-order moments of the class-conditional distributions, and achieve provable domain generalization with $d_s\mathrm{+}1$ training environments under the data model of \citet{risks-of-IRM}. Our second algorithm, ISR-Cov, further reduces the required number of training environments to $\cO(1)$ using the information of second-order moments. Notably, unlike IRM, our algorithms bypass non-convexity issues and enjoy global convergence guarantees. Empirically, our ISRs can obtain superior performance compared with IRM on synthetic benchmarks. In addition, on three real-world image and text datasets, we show that both ISRs can be used as simple yet effective post-processing methods to improve the worst-case accuracy of (pre-)trained models against spurious correlations and group shifts. The code is released at \url{https://github.com/Haoxiang-Wang/ISR}.
\end{abstract}
\vspace{-1.7em}
\section{Introduction}\label{sec:intro}
Domain generalization, i.e., out-of-distribution (OOD) generalization, aims to obtain models that can generalize to unseen (OOD) test domains after being trained on a limited number of training domains \citep{blanchard2011generalizing,wang2021generalizing,zhou2021domain,shen2021towards}. A series of works try to tackle this challenge by learning the so-called domain-invariant features (i.e., features whose distributions do not change across domains)~\citep{long2015learning,ganin2016domain,hoffman2018cycada,zhao2018adversarial,zhao2019learning}. On the other hand, Invariant Risk Minimization (IRM) \citep{IRM}, represents another approach that aims to learn features that induce invariant optimal predictors over training environments. Throughout this work, we shall use the term \emph{invariant features} to denote such features. There is a stream of follow-up works of IRM \citep{javed2020learning,REx,shi2020invariant,ahuja2020invariant,khezeli2021invariance}, which propose alternative objectives or extends IRM to different settings. 

Recently, some theoretical works demonstrate that IRM and its variants fail to generalize to unseen environments, or cannot outperform empirical risk minimization (ERM), in various simple data models \citep{risks-of-IRM,kamath2021does,ahuja2021empirical}. For instance, \citet{risks-of-IRM} considers a simple Gaussian linear data model such that the class-conditional distribution of \textit{invariant features} remains the same across domains, while that of \textit{spurious features} changes across domains. Intuitively, a successful domain generalization algorithm is expected to learn an \textit{optimal invariant predictor}, which relies on only the invariant features and is optimal over the invariant features. To remove the noise introduced by finite samples, these theoretical works generally assume that infinite samples are available per training environment to disregard finite-sample effects, and the main evaluation metric for domain generalization algorithms is the \textit{number of training environments} needed to learn an optimal invariant predictor -- this metric is also referred to as \textit{environment complexity} in the literature~\citep{chen2021iterative}. In the case of linear predictors, \citet{risks-of-IRM} shows that IRM and REx (an alternative objective of IRM proposed in \citep{REx}) need $E>d_s$ to learn optimal invariant predictors, where $E$ is the number of training environments, and $d_s$ is the dimension of spurious features. In the case of non-linear predictors, they both fail to learn invariant predictors. Notice that the $E > d_s$ condition of IRM can be interpreted as a \textit{linear environment complexity} (i.e., $O(d_s)$ complexity), which is also observed in other recent works \citep{kamath2021does,ahuja2021empirical,chen2021iterative}. 
% Notice that the $O(d_s)$ environment complexity is unrealistic for most real-world high-dimensional data, where usually have $E \ll d_s$. 

In this work, we propose a novel approach for domain generalization, Invariant-feature Subspace Recovery (ISR), that recovers the subspace spanned by the invariant features, and then fits predictors in this subspace. More concretely, we present two algorithms to realize this approach, ISR-Mean and ISR-Cov, which utilize the first-order and second-order moments (i.e., mean and covariance) of class-conditional distributions, respectively. Under the linear data model of \citet{risks-of-IRM} with linear predictors, we prove that a) ISR-Mean is guaranteed to learn the optimal invariant predictor with $E\geq d_s+1$ environment, matching the environment complexity of IRM, and b) ISR-Cov reduces the requirement to $E\geq 2$, achieving a constant $O(1)$ environment complexity. Notably, both of ISR-Mean and ISR-Cov require fewer assumptions on the data model than IRM, and they both enjoy global convergence guarantees, while IRM does not because of its non-convex formulation of the objective function. Notably, the ISRs are also more computationally efficient than algorithms such as IRM, since the computation of ISRs involves basically only the ERM with one additional call of an eigen-decomposition solver.

Empirically, we conduct studies on a set of challenging synthetic linear benchmarks designed by \cite{aubin2021linear} and a set of real-world datasets (two image datasets and one text dataset) used in \citet{sagawa2019distributionally}. Our empirical results on the synthetic benchmarks validate the claimed environment complexities, and also demonstrate its superior performance when compared with IRM and its variant. Since the real-world data are highly complex and non-linear, over which the ISR approach cannot be directly applied, we apply ISR on top of the features extracted by the hidden layers of trained neural nets as a post-processing procedure. Experiments show that ISR-Mean can consistently increase the worse-case accuracy of the trained models against spurious correlations and group shifts, and this includes models trained by ERM, reweighting and GroupDRO~\citep{sagawa2019distributionally}.\looseness=-1

\vspace{-.3em}
\section{Related Work}\label{sec:related-works}
\vspace{-.3em}

\textbf{Domain Generalization.}~Domain generalization (DG), also known as OOD generalization, aims at leveraging the labeled data from a limited number of training environments to improve the performance of learning models in unseen test environments~\citep{blanchard2011generalizing}. The simplest approach for DG is empirical risk minimization \cite{vapnik1992principles}, which minimizes the sum of empirical risks over all training environments. Distributionally robust optimization is another approach \citep{sagawa2019distributionally,volpi2018generalizing}, which optimizes models over a worst-case distribution that is perturbed around the original distribution. Besides, there are two popular approaches, domain-invariant representation learning and invariant risk minimization, which we will discuss in detail below. In addition to algorithms, there are works that propose theoretical frameworks for DG \citep{zhang2021quantifying,ye2021towards}, or empirically examine DG algorithms over various benchmarks \citep{gulrajani2021in,koh2021wilds,wiles2021fine}. 
Notably, some recent works consider DG with temporarily shifted environments \citep{koh2021wilds,ye2022future}, which is a novel and challenging setting. Besides DG, there are other learning paradigms that involve multiple environments, such as multi-task learning \citep{caruana1997multitask,wang2021bridging} and meta-learning \citep{finn2017model,wang2022global}, which do not aim at generalization to OOD environments.

\textbf{Domain-Invariant Representation Learning.}~Domain-Invariant representation learning is a learning paradigm widely applied in various tasks. In particular, in domain adaptation (DA), many works aim to learn a representation of data that has an invariant distribution over the source and target domains, adopting methods including adversarial training \citep{ganin2016domain,tzeng2017adversarial,zhao2018adversarial} and distribution matching \citep{ben2007analysis,long2015learning,sun2016deep}. The domain-invariant representation approach for DA enjoys theoretical guarantees~\citep{ben2010theory}, but it is also pointed out that issues such as conditional shift should be carefully addressed~\citep{zhao2019domain}. In domain generalization~\citep{blanchard2011generalizing}, since there is no test data (even unlabelled ones) available, models are optimized to learn representations invariant over training environments~\citep{albuquerque2020generalizing,chen2021iterative}. Notice that many domain-invariant representation learning methods for DA can be easily applied to DG as well~\citep{gulrajani2021in}.

\textbf{Invariant Risk Minimization.}~\citet{IRM} proposes invariant risk minimization (IRM) that aims to learn invariant predictors over training environments by optimizing a highly non-convex bi-level objective. The authors also reduce the optimization difficulty of IRM by proposing a practical version, IRMv1, with a penalty regularized objective instead of a bi-level one. Alternatives of IRM have also been studied \citep{ahuja2020invariant,li2021invariant}. However, \citet{risks-of-IRM,kamath2021does,ahuja2021empirical} theoretically show that these algorithms fail even in simple data models.

\begin{figure}[h]
\begin{center}
% \vskip -0.1in
% \vspace{-1em}
\centerline{\includegraphics[width=.5\columnwidth]{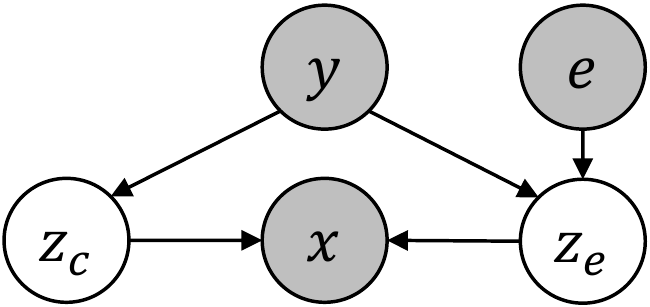}}
% \vspace{-.4em}
% \vskip -0.1in
\caption{The causal graph of the data model in \citet{risks-of-IRM}. Shading represents that the variable is observed.
}
\label{fig:cgraph}
\end{center}
\vspace{-1.1em}
\end{figure}
% \vspace{-0.3em}
\section{Problem Setup}\label{sec:setup}
% \vspace{-0.3em}
\textbf{Notations}~ Each labeled example can be represented as a $(x,y,e)$ tuple, where $x\in \bR^d$ is the input, $y\in \{\pm 1\}$ is the label, and $e\in \mathbb{Z}_+$ is the index of the environment that provides $(x,y)$. In addition, we assume $x$ is generated by a latent feature $z\in \bR^d$, which generates $x$ and is correlated with $y$ and $e$  (e.g., see the example in Fig.~\ref{fig:cgraph}). Besides, we use $X,Y,\mathscr E, Z$ to refer to random variables w.r.t. $x,y,e,z$.

\textbf{Data Model}~ In this paper, we adopt the linear Gaussian data model of \citet{risks-of-IRM}, which assumes that training data are drawn from $E$ training environments, $\mathcal E = \{1,..., E\}$. For arbitrary training environment $e\in \mathcal E$, each sample in this environment is generated by the following mechanism (see Fig. \ref{fig:cgraph} for an illustration): first, a label $y\in \{\pm 1\}$ is sampled,
% \vspace{-1em}
\begin{align} \label{eq:def-y}
    y & = \begin{cases}
    1, & \text{with probability } \eta\\
    -1, & \text{otherwise}
    \end{cases}
\end{align}
Then, both invariant latent features $z_c$ and spurious latent features $z_e$ of this sample are drawn from the following Gaussian distributions:
\begin{align}\label{eq:def-latent-features}
    z_c \sim \mathcal{N}(y\mu_c, \sigma_c^2 I) \in \mathbb{R}^{d_c}, 
    z_e \sim \mathcal{N}(y  \mu_e, \sigma_e^2 I) \in \mathbb{R}^{d_s}  
\end{align}
where $\mu_c \in \mathbb{R}^{d_c}, \mu_e \in \mathbb{R}^{d_s}$ and $\sigma_c,\sigma_e \in \mathbb{R}_+$. The constants $d_c$ and $d_s$ refer to the dimension of invariant features and spurious features, respectively. The total number of feature attributes is then $d=d_c+d_s$. Notice that $\mu_c,\sigma_c$ are invariant across environments, while $\mu_e,\sigma_e$ are dependent on the environment index $e$. Following \citet{risks-of-IRM}, we name $\{\mu_e\}$ and $\{\sigma_e\}$ as \textit{environmental} means and variances.

\citet{risks-of-IRM} adopts a mild non-degeneracy assumption\footnote{It was stated as (9) in \citet{risks-of-IRM}.} on the environmental mean from the IRM paper \citep{IRM}, stated as Assumption \ref{assum:non-degenerate-mean} below. In addition, the authors also make another non-degeneracy assumption\footnote{It is stated as Eq. (10) in \citet{risks-of-IRM}, which is a sufficient (not necessary) condition for our Assumption \ref{assum:non-degenerate-cov}.} on the environmental variances, which we relax to the following Assumption \ref{assum:non-degenerate-cov}.
\begin{assumption}\label{assum:non-degenerate-mean} For the set of environmental means, $\{\mu_e\}_{e=1}^E$, we assume that each element of the set cannot be expressed as an affine combination of the rest elements.
\end{assumption}
\begin{assumption}\label{assum:non-degenerate-cov}
Assume there exists a pair of distinct training environments $e,e'\in[E]$ such that $\sigma_e\neq \sigma_{e'}$.
\end{assumption}
With the latent feature $z$ as a concatenation of $z_c$ and $z_e$, the observed sample $x$ is generated by a linear transformation on this latent feature. For simplicity, we consider that $x$ has the same dimension as $z$.
\begin{align}\label{eq:def-linear-transform}
    z = \begin{bmatrix}
    z_c\\
    z_e
    \end{bmatrix} \in \bR^{d}, \quad 
    x = R z = A z_c + B z_e \in \bR^d
\end{align}
where $d= d_c + d_s$, and $A = \bR^{d \times d_c}, B = \bR^{d \times d_s}$ are fixed transformation matrices with concatenation as $R = [A,B]\in \bR^{d\times d}$. Then, each observed sample $x$ is effectively a sample drawn from 
\begin{align}\label{eq:transformed-gaussian}
    \mathcal N(y(A \mu_c + B \mu_e), \sigma_c^2 AA^\T + \sigma_e^2 BB^\T)
\end{align}
The following assumption is also imposed on the transformation matrix in~\citet{risks-of-IRM}:
\begin{assumption}\label{assum:full-rank-transform} $R$ is injective.
\end{assumption}
\vspace{-.3em}
Since $R\in \bR^{d \times d}$, Assumption \ref{assum:full-rank-transform} leads to the fact $\mathrm{rank}(R) = d$, indicating that $R$ is full-rank.

Denote the data of any training domain $e$ as $\mathcal D_e$. During training, learners have access to the environment index $e$ for each training sample, i.e., learners observe samples in the form of $(x,y,e)$.

\textbf{Optimal Invariant Predictors}~ The quest of IRM is to find the optimal invariant predictors, i.e., classifiers that use only invariant features and are optimal w.r.t. invariant features over the training data. In the data model of \citet{risks-of-IRM}, because of the linear nature of the data generation process, the optimal invariant predictors are contained in the linear function class. Since the task of consideration is binary classification, \citet{risks-of-IRM} chooses the logistic loss as the loss function for optimization\footnote{\citet{risks-of-IRM} proves that logistic loss over linear models can attain Bayes optimal classifiers in this data model.}, which we also adopt in this work. Then, the goal of domain generalization in this data model is to learn a linear featurizer (feature extractor) $\Phi$ and a linear classifier $ \beta$ that minimizes the risk (population loss) on any unseen environment $e$ with data distribution $p_e$ satisfying Assumptions~\eqref{eq:def-y}-\eqref{eq:def-linear-transform}:
\begin{align}
    \mathcal{R}^{e}(\Phi, \beta):=\mathbb{E}_{(x, y) \sim p^{e}}\left[\ell\left(w^{\T} \Phi(x) + b ,~ y\right)\right]
\end{align}
where $\ell$ is the logistic loss function, and $\beta=(w,b)$ with weight $w$ and bias $b$.

To be complete, we present the optimal invariant predictor derived by \citet{risks-of-IRM} as follows.
\vspace{-.4em}
\begin{proposition}[Optimal Invariant Predictor]\label{prop:optimal-inv-pred}
Under the data model considered in Eq. \eqref{eq:def-y}-\eqref{eq:def-linear-transform}, the optimal invariant predictor $h^*$ w.r.t. logistic loss is unique, which can be expressed as a composition of i) a featurizer $\Phi^*$ that recovers the invariant features and ii) the classifier $\beta^*=(w^*,b^*)$ that is optimal w.r.t. the extracted features:
\begin{align}
    h^*(x) &= {w^*}^\T \Phi^*(x) + b^*\label{eq:optimal-inv-pred}\\
    \Phi^*(x)&\coloneqq
    \begin{bmatrix}
    I_{d_c} & 0\\
    0 & 0
    \end{bmatrix} R^{-1} x= 
    \begin{bmatrix}
    z_c\\
    0
    \end{bmatrix}\in \bR^{d \times d}\\
    w^*&\coloneqq 
    \begin{bmatrix}
    2 \mu_c / \sigma_c^2\\
    0
    \end{bmatrix}\in \bR^{d}, \quad 
    b^* \coloneqq \log \frac{\eta}{1-\eta} \in \bR
\end{align}
\end{proposition}
Notice that even though the optimal invariant predictor $h^*$ is unique, its components (the featurizer and classifier) are only unique up to invertible transformations. For instance, $( {w^*}^\T U^{-1})(U \Phi) = {w^*}^\T \Phi $ for any invertible $U\in \bR^{d\times d}$.

\textbf{Invariant Risk Minimization}~ IRM optimizes a bi-level objective over a featurizer $\Phi$ and a classifier $\beta$,
\begin{align}\label{eq:IRM}
    \mathrm{IRM:}~~&\min_{\Phi,\beta} \sum_{e\in [E]} \mathcal{R}^{e}(\Phi, \beta) \\
    &\mathrm{s.t.}~ \beta \in \argmin_{\beta}\mathcal{R}^e(\Phi, \beta) ~~\forall e\in[E]\nonumber
\end{align}
This objective is non-convex and difficult to optimize. Thus, \citet{IRM} proposes a Langrangian form to find an approximate solution, 
\begin{align}\label{eq:IRMv1}
    \mathrm{IRMv1:}~ \min _{\Phi, \hat{\beta}} \sum_{e \in [E]}\mathcal{R}^{e}(\Phi, \hat{\beta})+\lambda\left\|\nabla_{\hat{\beta}} \mathcal{R}^{e}(\Phi, \hat{\beta})\right\|_{2}^{2}
\end{align}
where $\lambda > 0$ controls the regularization strength. Notice that the IRMv1$\eqref{eq:IRMv1}$ is still non-convex, and it becomes equivalent to the original IRM \eqref{eq:IRM} as $\lambda \rightarrow \infty$.

\textbf{Environment Complexity}~~To study the dependency of domain generalization algorithms on environments, recent theoretical works \citep{risks-of-IRM,kamath2021does,ahuja2021invariance,chen2021iterative} consider the ideal setting of infinite data per training environment to remove the finite-sample effects. In this infinite-sample setting, a core measure of domain generalization algorithms is \textit{environment complexity}: the number of training environments needed to learn an invariant optimal predictor. For this data model, \citet{risks-of-IRM} proves that with the linear $\Phi$ and $\beta$, the environment complexity of IRM is $d_s+1$, assuming the highly non-convex objective \eqref{eq:IRM} is optimized to reach the global optimum. This linear environment complexity (i.e., $O(d_s)$) of IRM is also proved in \citep{kamath2021does,ahuja2021invariance} under different simple data models.

\setlength{\textfloatsep}{10pt}
\begin{algorithm}[tb]
\caption{ISR-Mean}\label{algo:isr-mean}
\begin{algorithmic}
\STATE {\bfseries Input:} Data of all training environments, $\{\mathcal D_e\}_{e\in [E]}$.
\FOR{$e = 1,2,\dots,E$}     
    \STATE Estimate the sample mean of $\{x|(x,y)\in \mathcal D_e, y=1\}$ as $\bar x_e\in \bR^{d}$
\ENDFOR
\STATE \textbf{I.} Construct a matrix $\mathcal M \in \bR^{E \times d}$ with the $e$-th row as $\bar{x}_e^\T$ for $e\in [E]$
\STATE \textbf{II.} Apply PCA to $\mathcal M$ to obtain eigenvectors $\{P_1,...,P_d\}$ with eigenvalues $\{\lambda_1,...,\lambda_d\}$
\STATE \textbf{III.} Stack $d_c$ eigenvectors with the lowest eigenvalues to obtain a transformation matrix $P'\in \bR^{d_c\times d}$
\STATE \textbf{IV.} Fit a linear classifier (with $w\in \bR^{d_c}$, $b\in \bR$) by ERM over all training data with transformation $x\mapsto P'x$
\STATE Obtain a predictor $f(x) =  w^\T P' x + b$
\end{algorithmic}
\end{algorithm}

\section{Invariant-Feature Subspace Recovery}\label{sec:algo}

In this section, we introduce two algorithms, ISR-Mean and ISR-Cov, which recover the invariant-feature subspace with the first-order and second-order moments of 
class-conditional data distributions, respectively. 
\subsection{ISR-Mean} \label{sec:isr-mean}
Algorithm \ref{algo:isr-mean} shows the pseudo-code of ISR-Mean, and we explain its four main steps in detail below. In the setup of \cref{sec:setup}, ISR-Mean enjoys a linear environment complexity that matches that of IRM, while requiring fewer assumptions (no need for Assumption \ref{assum:non-degenerate-cov}).

\textbf{I. Estimate Sample Means across Environments}~
In any training environment $e$, each observed sample $x\in \bR^d$ is effectively drawn i.i.d. from $\mathcal N (y (A \mu_c + B \mu_e), A A^\T \sigma_c^2 + B B^\T \sigma_e^2)$, as stated in \eqref{eq:transformed-gaussian}. In the infinite-sample setting considered in \cref{sec:setup}, the mean of the positive-class data in environment $e$ can be expressed as $\E[X|Y=1, \mathscr E =e]$, which is exactly the value of $\bar{x}_e$ in \cref{algo:isr-mean}. Thus, we know $\bar x_e$ satisfies $\bar{x}_e = A \mu_c + B \mu_e$, and the matrix $\mathcal M$ can be expressed as
\vspace{-1em}
\begin{align}\label{eq:def-M}
    \mathcal M \mathrm{\coloneqq} \begin{bmatrix}
    \bar{x}_1^\T\\
    \vdots\\
    \bar{x}_E^\T
    \end{bmatrix} \mathrm{=} \begin{bmatrix}
    \mu_c^\T A^\T \mathrm{+} \mu_1^\T B^\T\\
    \vdots\\
    \mu_c^\T A^\T \mathrm{+} \mu_E^\T B^\T\end{bmatrix}
    \mathrm{=}{\overbrace{\begin{bmatrix}
    \mu_c^\T ~~\mu_1^\T\\
    \vdots~~~~~\vdots\\
    \mu_c^\T ~~ \mu_E^\T
    \end{bmatrix}}^{\mathcal U^\T \coloneqq}} R^\T
\end{align}

\begin{figure}[t!]
\vspace{-.4em}
\begin{center}
\centerline{\includegraphics[width=.9\columnwidth]{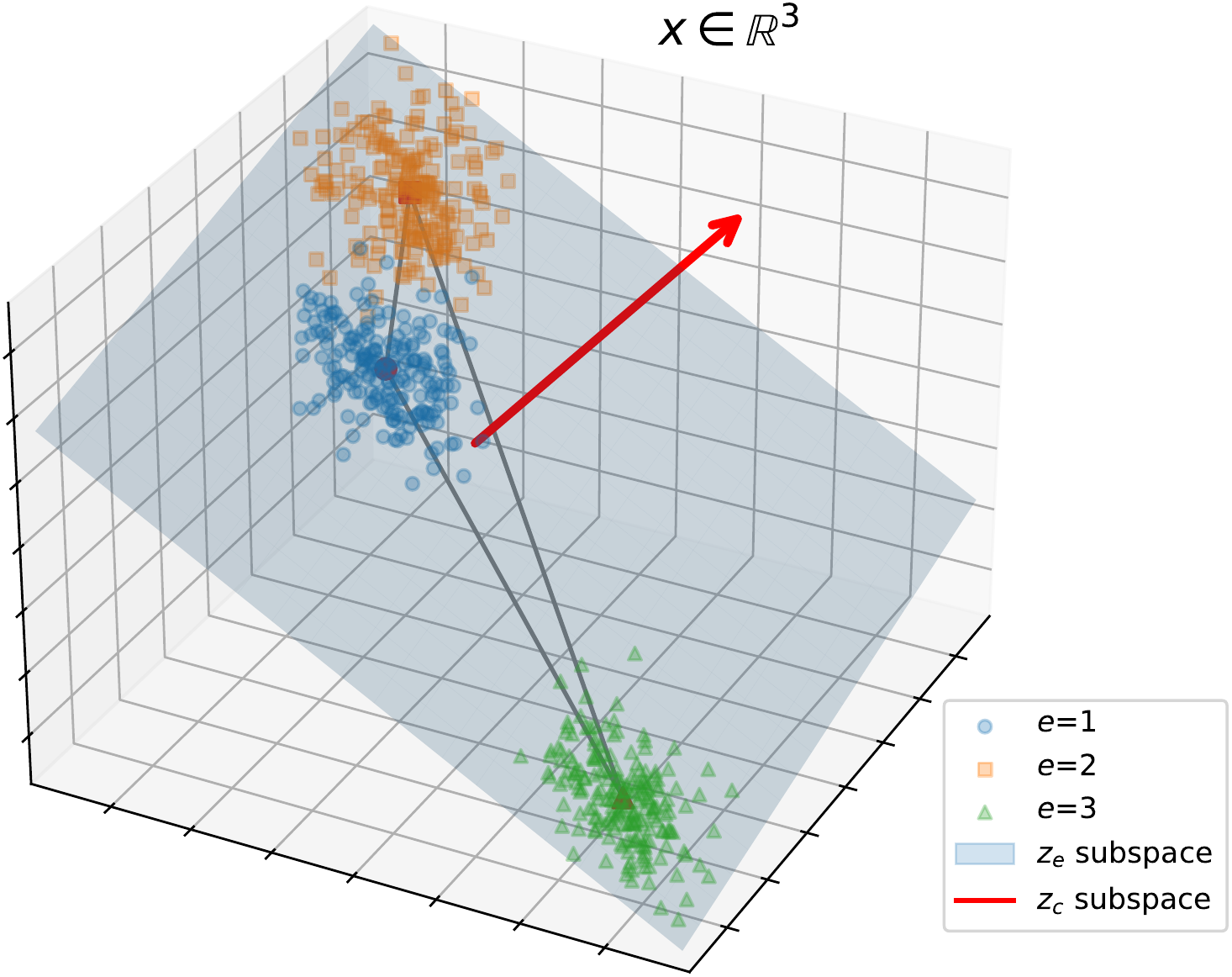}}
\vspace{-1em}
\caption{An example for ISR-Mean with $d_c\mathrm{=}1$, $d_s\mathrm{=}2$, $E\mathrm{=}3$. In this $\bR^3$ input space, the blue 2D plane is determined by sample means of positive-class samples of the $3$ training environments. 
}
\label{fig:ISR-mean-demo}
\end{center}
\vspace{-1.5em}
\end{figure}

\textbf{II. PCA on $\mathcal M$.}~
In this step, we apply principal component analysis (PCA) \citep{pearson1901PCA} to the matrix $\mathcal M$. First, PCA performs mean-substraction on $\mathcal M$ to shift the sample mean of each column to zero, and we denote the shifted matrix as $\wt{\mathcal M}$. Then, PCA eigen-decompose $\wh \Sigma_\cM \coloneqq \frac{1}{E} \wt{\cM}^\T \wt{\cM}$, the sample covariance matrix of $\wt\cM$, such that $\wh \Sigma_\cM = P^\T S P$, where $P = [P_1,\dots,P_d]\in \bR^{d\times d}$ is a stack of eigenvectors $\{P_i\}_{i\in d}^d$, and $S\in \bR^{d \times d}$ is a diagonal square matrix with diagonal entries as eigenvalues $\{\lambda_i\}_{i=1}^d$ of $\wt \Sigma_{\cM}$. We consider the eigenvalues $\{\lambda_i\}_{i=1}^d$ are sorted in ascending order.

\paragraph{III. Recover the Invariant-Feature Subspace}
As we shall formally prove in \cref{thm:isr-mean}, in the infinite-sample setting, a) the eigenvalues $\{\lambda_i\}_{i\in 1}^d$ should exhibit a ``phase transition'' phenomenon such that the first $d_c$ eigenvalues all are \textit{zeros} while the rest are all \textit{positive}, b) the $d_c$ eigenvectors corresponding to zero eigenvalues, $\{P_1,\dots,P_{d_c}\}$, are guaranteed to recover the $d_c$-dimensional subspace spanned by invariant latent feature dimensions, i.e., the subspace of $z_c$ defined in \eqref{eq:def-latent-features}. We stack these eigenvectors as a matrix $P'$ \looseness=-1
\begin{align}
    P' \coloneqq [P_1,\dots, P_{d_c}]^\T \in \bR^{d_c \times d}
\end{align}
% \vspace{-1.8em}

\textbf{IV. Train a Classifier in the Invariant-Feature Subspace}~
In this final step, we just transform all training data by the transformation $x\mapsto P'x$, and fit a linear classifier with ERM to the transformed data to obtain an predictor,
\begin{align}\label{eq:ISR-Mean-predictor}
    f(x) = w^\T P' x + b
\end{align}
which is guaranteed to be the optimal invariant predictor $h^*$ defined in \cref{prop:optimal-inv-pred}, i.e., $f\equiv h^*$.

\textbf{Global Convergence Guarantee}~ ISR-Mean is guaranteed to converge to a global optimum since a) the step I and III are optimization-free, b) PCA can be efficiently optimized to global the optimum by various methods \cite{arora2012stochastic,vu2013fantope,hauser2018pca,eftekhari2020principal}, c) the ERM objective of linear classifiers with logistic loss is convex, enjoying global convergence.

\textbf{Geometric Interpretation.}~ We provide an geometric interpretation of ISR-Mean with a 3D example in Fig.~\ref{fig:ISR-mean-demo}, where $d_c\mathrm{=}1$, $d_s\mathrm{=}2$, $E\mathrm{=}3$. For each environment $e$, the sample mean of its positive-class data, $\bar{x}_e$, must lie in a $d_s$-dimensional spurious-feature subspace in the infinite-sample setting, as proved by \cref{thm:isr-mean}. ISR-Mean aims to identify this spurious-feature subspace, and take its tangent subspace as the invariant-feature subspace.

\textbf{Linear Environment Complexity}~ 
In the infinite-sample setting, we prove below that with more than $d_s$ training environments, ISR-Mean is guaranteed to learn the invariant optimal predictor (\cref{thm:isr-mean}). Notice that even though this linear environment complexity is identical to that of IRM (proved in Theorem 5.1 of \citet{risks-of-IRM}), our ISR-Mean has two additional advantages: (a) Unlike IRM, ISR-Mean does not require any assumption on the covariance\footnote{IRM needs a covariance assumption stronger than our Assumption \ref{assum:non-degenerate-cov}, as pointed out in Sec. \ref{sec:setup}.} such as Assumption \ref{assum:non-degenerate-cov}; (b) ISR-Mean enjoys the global convergence guarantee, while IRM does not due to its non-convex formulation. The proof is in \cref{supp:proof}.

\begin{theorem}[ISR-Mean]\label{thm:isr-mean}
Suppose $E > d_s$ and the data size of each environment is infinite, i.e., $|\mathcal D_e| \mathrm{\rightarrow} \infty$ for $e\mathrm{=}1,\dots,E$. For PCA on the $\cM$ defined in \eqref{eq:def-M}, the obtained eigenvectors $\{P_1,\dots, P_d\}$ with corresponding ascendingly ordered eigenvalues $\{\lambda_1,\dots, \lambda_d\}$ satisfy
\begin{align*}
    \forall 1\leq i \leq d_c, ~\lambda_i = 0 \quad \text{and} \quad  \forall d_c < i \leq d, ~\lambda_i > 0
\end{align*}
The eigenvectors corresponding to these zero eigenvalues, i.e., $\{P_1,\dots, P_{d_c}\}$, can recover the subspace spanned by the invariant latent feature dimensions, i.e.,
\begin{align}\label{eq:thm:ISR-Mean:span-equality}
\mathrm{Span}(\{P_1^\T R,\dots, P_{d_c}^\T R\}) = \mathrm{Span}(\{\mathbf{\hat 1},\dots, \mathbf{\hat d_c}\})
\end{align}
where $\mathbf{\hat i}$ is the unit-vector along the $i$-th coordinate in the latent feature space for $i=1,\dots, d$. Then, the classifier $f$ fitted with ERM to training data transformed by $x\mapsto [P_1,\dots,P_{d_c}]^\T x$ is guaranteed to be the invariant optimal predictor, i.e., $f = h^*$, where $h^*$ is defined in \eqref{eq:optimal-inv-pred}.
\end{theorem} 
\vspace{-.5em}
\subsection{ISR-Cov} \label{sec:isr-cov}
\vspace{-.5em}
The pseudo-code of ISR-Cov\footnote{In the final stage of this paper preparation, we notice a concurrent work, the v2 of \citet{chen2021iterative} (uploaded to arXiv on Nov 22, 2021), appends a new algorithm in its Appendix C that is similar to our ISR-Cov, under data model assumptions stricter than ours. That algorithm does not exist in their v1.} is presented in Algorithm \ref{algo:isr-cov}, with a detailed explanation below.
In the setup of \cref{sec:setup}, ISR-Cov attains an $O(1)$ environment complexity, the optimal complexity any algorithm can hope for, while requiring fewer assumptions than IRM (no need for Assumption \ref{assum:non-degenerate-mean}).

\textbf{I. Estimate and Select Sample Covariances across Environments}~
As \eqref{eq:transformed-gaussian} indicates, in any environment $e$, each observed sample $x\in \bR^d$ with $y=1$ is effectively drawn i.i.d. from $\mathcal N (A \mu_c + B \mu_e, A A^\T \sigma_c^2 + B B^\T \sigma_e^2)$. Thus, the covariance of the positive-class data in environment $e$ can be expressed as $\mathrm{Cov}[X | Y=1, \mathscr{E}=e] = A A^\T \sigma_c^2 + B B^\T \sigma_e^2$, which is the value that $\Sigma_e$ in the step I of \cref{algo:isr-cov} estimates. The estimation is exact in the infinite-sample setting of consideration, so we have $\Sigma_e = A A^\T \sigma_c^2 + B B^\T \sigma_e^2$. Assumption \ref{assum:non-degenerate-cov} guarantees that we can select a pair of environments $e_1,e_2$ with $\Sigma_1 \neq \Sigma_2$. Then, we have
\begin{align}\label{eq:delta-sigma-expression}
\Delta \Sigma \coloneqq \Sigma_{e_1} - \Sigma_{e_2} = (\sigma_{e_1}^2 - \sigma_{e_2}^2)BB^\T \in \bR^{d \times d}
\end{align}

\textbf{II. Eigen-decompose $\Delta \Sigma$} Similar to the step II of \cref{algo:isr-mean} explained in Sec. \ref{sec:isr-mean}, we eigen-decompose $\Delta\Sigma$ to obtain eigenvectors $\{P_i\}_{i\in d}^d$ corresponding to eigenvalues $\{\lambda_i\}_{i=1}^d$. We consider the eigenvalues are sorted in ascending order by their \textit{absolute values}.

\textbf{III. Recover the Invariant-Feature Subspace}~
As we shall formally prove in \cref{thm:isr-cov}, in the infinite-sample setting, a) the eigenvalues $\{\lambda_i\}_{i\in 1}^d$ should exhibit a ``phase transition'' phenomenon such that the first $d_c$ eigenvalues all are \textit{zeros} while the rest are all \textit{non-zero}, b) the $d_c$ eigenvectors corresponding to zero eigenvalues, $\{P_1,\dots,P_{d_c}\}$, are guaranteed to recover the $d_c$-dimensional invariant-feature subspace. We stack these eigenvectors as a matrix $P'$
\begin{align}\label{eq:trans-matrix-isr-cov}
    P' \coloneqq [P_1,\dots, P_{d_c}]^\T \in \bR^{d_c \times d}
\end{align}

\textbf{IV. Train a Classifier in the Invariant-Feature Subspace}~
This final step is the same as the step IV of Algorithm \ref{algo:isr-mean} described in Sec. \ref{sec:isr-mean}.

\paragraph{Global Convergence}
Applying the same argument in Sec.~\ref{sec:isr-mean}, it is clear that ISR-Cov also enjoys the global convergence guarantee: the eigen-decomposition and ERM can both be globally optimized.

\textbf{Improving the Robustness of ISR-Cov} In practice with finite data, Algorithm \ref{algo:isr-cov} may be non-robust as $\sigma_e$ and $\sigma_{e'}$ become close to each other. The noise due to finite samples could obfuscate the non-zero eigenvalues so that they are indistinguishable from the zero eigenvalues. To mitigate such issues, we propose a robust version of ISR-Cov that utilizes more pairs of the given environments. Briefly speaking, the robust version is to \textbf{a)} run the step I to III of ISR-Cov over $N \leq \binom E 2$ pairs of environments with distinct sample covariances, leading to $N$ $d_c$-dimensional subspaces obtained through Algorithm \ref{algo:isr-cov}, \textbf{b)} we find the stable $d_c$-dimensional subspace, and use it as the invariant-feature subspace that we train the following classifier. Specifically, we achieve b) by computing the flag-mean \citep{marrinan2014finding} over the set of $P'$ (defined in \eqref{eq:trans-matrix-isr-cov}) obtained from the $N$ selected pairs of environments. Compared with the original ISR-Cov, this robust version makes use of training data more efficiently (e.g., it uses more than one pair of training environments) and is more robust in the finite-data case. We implement this robust version of ISR-Cov in our experiments in Sec.~\ref{sec:exp}.

\setlength{\textfloatsep}{5pt}
\begin{algorithm}[t!]
\caption{ISR-Cov}\label{algo:isr-cov}
\begin{algorithmic}
\STATE {\bfseries Input:} Data of all training environments, $\{\mathcal D_e\}_{e\in [E]}$.
\FOR{$e = 1,2,\dots,E$}     
    \STATE Estimate the sample covriance of $\{x|(x,y)\in \mathcal D_e, y=1\}$ as $\Sigma_e \in \bR^{d \times d}$
\ENDFOR
\STATE \textbf{I.} Select a pair of environments $e_1,e_2$ such that $\Sigma_1 \neq \Sigma_2$, and compute their difference, $\Delta \Sigma \coloneqq \Sigma_{e_1} - \Sigma_{e_2}$
\STATE \textbf{II.} Eigen-decompose $\Delta\Sigma$ to obtain eigenvectors $\{P_1,...,P_d\}$  with eigenvalues $\{\lambda_1,...,\lambda_d\}$
\STATE \textbf{III.} Stack $d_c$ eigenvectors of eigenvalues with lowest absolute values to obtain a matrix $P'\in \bR^{d_c\times d}$
\STATE \textbf{IV.} Fit a linear classifier (with $w\in \bR^{d_c}$, $b\in \bR$) by ERM over all training data with transformation $x\mapsto P'x$
\STATE Obtain a predictor $f(x) =  w^\T P' x + b$
\end{algorithmic}
\end{algorithm}

\begin{figure}[t!]
\begin{center}
\centerline{\includegraphics[width=.9\columnwidth]{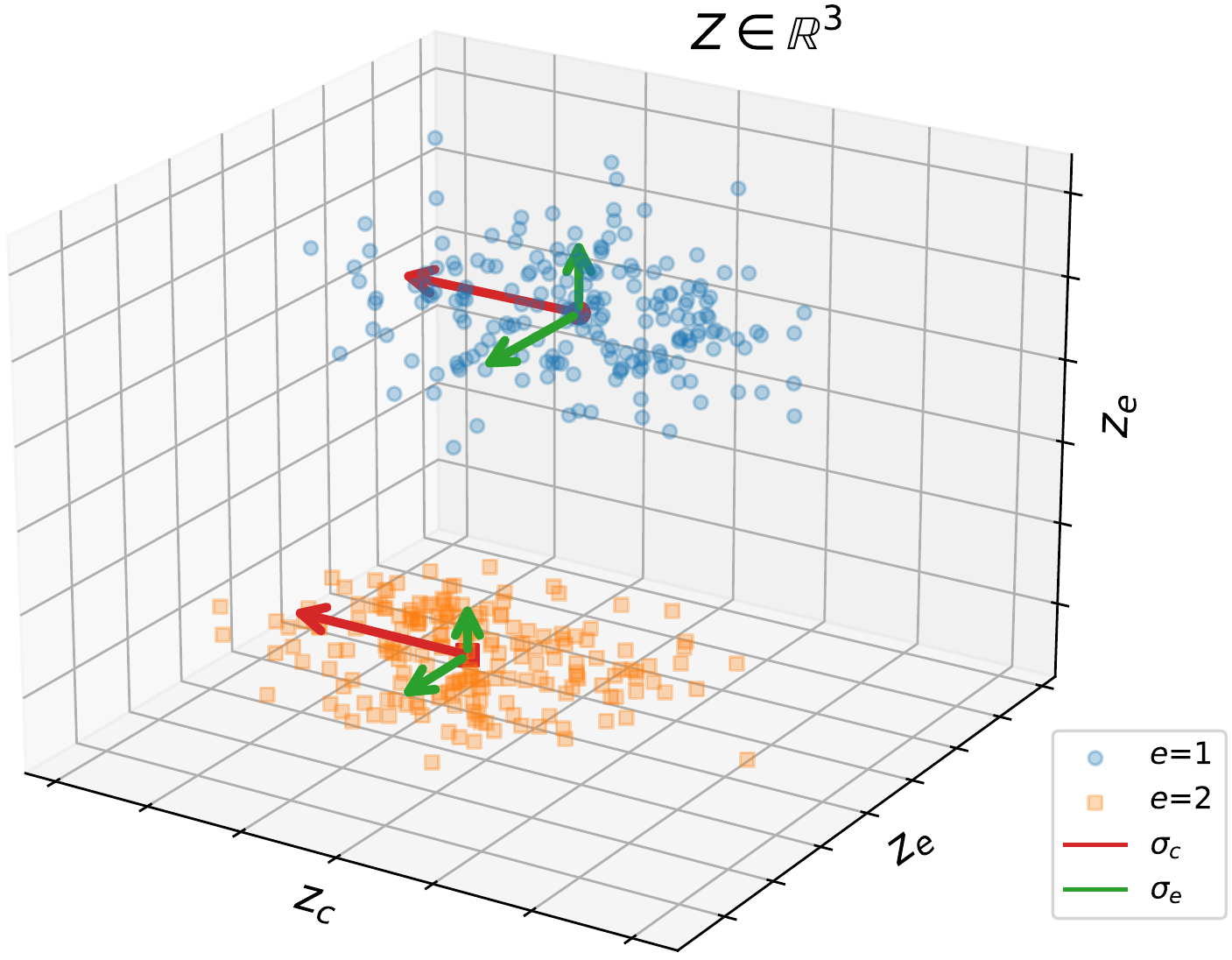}}
\vskip -0.1in
\caption{An example for ISR-Cov, where $d_c\mathrm{=}1$, $d_s\mathrm{=}2$, $E\mathrm{=}2$. In this latent feature space of $z\in\bR^3$, there is one dimension of $z_c$ and the rest two of $z_e$. 
}
\label{fig:ISR-cov-demo}
\end{center}
\vskip -0.15in
\end{figure}

\vspace{-.6em}
\paragraph{Geometric Interpretation}
We provide an geometric interpretation of ISR-Cov with a 3D example in Fig.~\ref{fig:ISR-cov-demo}, where $d_c\mathrm{=}1$, $d_s\mathrm{=}2$, $E\mathrm{=}2$. For either environment $e\in\{1,2\}$, the covariance of its class-conditional latent-feature distribution, $\begin{bmatrix}
\sigma_c^2 I_{d_c} & 0\\
0 & \sigma_e^2 I_{d_s}
\end{bmatrix}$, is \textit{anisotropic}: the variance $\sigma_c$ along invariant-feature dimensions is constant, while $\sigma_e$ along the spurious-feature dimensions is various across $e\in \{1,2\}$ (ensured by Assumption \ref{assum:non-degenerate-cov}). Though the transformation $R$ is applied to latent features, ISR-Cov still can identify the subspace spanned by invariant-feature dimensions in the latent-feature space by utilizing this anisotropy property. 

\textbf{$\cO(1)$ Environment Complexity}~
In the infinite-sample setting, we prove below that as long as there are at least two training environment that satisfies Assumption \ref{assum:non-degenerate-cov} and \ref{assum:full-rank-transform}, ISR-Cov is guaranteed to learn the invariant optimal predictor. This is the minimal possible environment complexity, since spurious and invariant features are indistinguishable with only one environment. Notably, unlike IRM, a) ISR-Cov does not require Assumption \ref{assum:non-degenerate-mean}, and b) ISR-Cov has a global convergence guarantee. The proof is in \cref{supp:proof}.

\begin{theorem}[ISR-Cov]\label{thm:isr-cov} Suppose $E\geq 2$ and the data size of each environment is infinite, i.e., $|\mathcal D_e| \mathrm{\rightarrow} \infty$ for $e\mathrm{=}1,\dots,E$. Eigen-decomposing $\Delta \Sigma$ defined in \eqref{eq:delta-sigma-expression}, the obtained eigenvectors $\{P_1,\dots, P_d\}$ with corresponding eigenvalues $\{\lambda_1,\dots, \lambda_d\}$ (ascendingly ordered by absolute values) satisfy
\begin{align*}
    \forall 1\leq i \leq d_c, ~\lambda_i = 0 \quad \text{and} \quad  \forall d_c < i \leq d, ~\lambda_i \neq 0
\end{align*}
The eigenvectors corresponding to these zero eigenvalues, i.e., $\{P_1,\dots, P_{d_c}\}$, can recover the subspace spanned by the invariant latent feature dimensions, i.e.,
\begin{align}\label{eq:thm:ISR-Cov:span-equality}
\mathrm{Span}(\{P_1^\T R,\dots, P_{d_c}^\T R\}) = \mathrm{Span}(\{\mathbf{\hat 1},\dots, \mathbf{\hat d_c}\})
\end{align}
where $\mathbf{\hat i}$ is the unit-vector along the $i$-th coordinate in the latent feature space for $i=1,\dots, d$. Then, the classifier $f$ fitted with ERM to training data transformed by $x\mapsto [P_1,\dots,P_{d_c}]^\T x$ is guaranteed be the invariant optimal predictor, i.e., $f = h^*$, where $h^*$ is defined in \eqref{eq:optimal-inv-pred}.
\end{theorem}

\section{Experiments}\label{sec:exp}

\begin{figure*}[t!]
\begin{center}
\vspace{-0.2em}
\centerline{\includegraphics[width=1\linewidth]{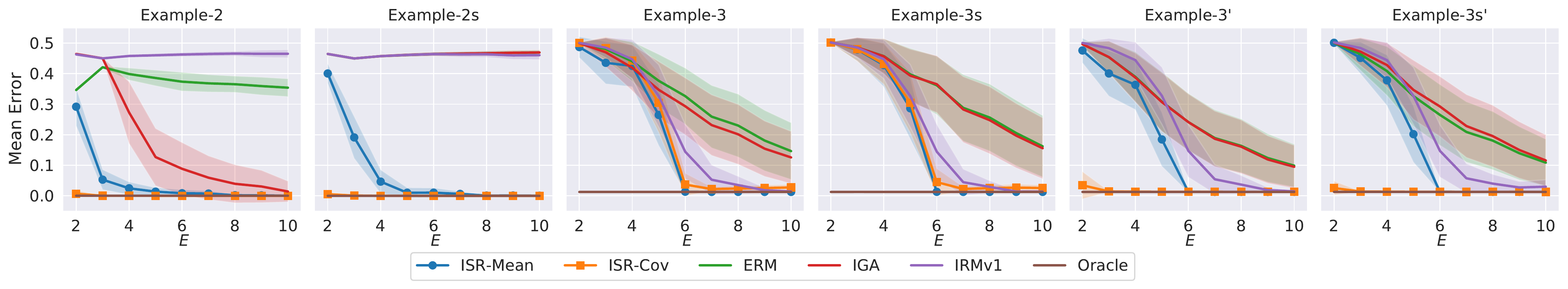}}
\vskip -0.15in
\caption{Test results on Linear Unit-Tests (first 4 plots) and its variants (last 2 plots), where $d_c = 5, d_s=5$, and $E=2,\dots, 10$.
}
\label{fig:linear-unit-tests}
\end{center}
\vskip -0.3in
\end{figure*}

We conduct experiments on both synthetic and real datasets to examine our proposed algorithms.

\subsection{Synthetic Datasets: Linear Unit-Tests}  \label{sec:lut}
We adopt a set of synthetic domain generalization benchmarks, Linear Unit-Tests \citep{aubin2021linear}, which is proposed by authors of IRM and is used in multiple recent works \citep{koyama2020out,khezeli2021invariance,du2021beyond}.
Specifically, we take four classification benchmarks\footnote{The rest are regression benchmarks, which we do not study.} from the Linear Unit-Tests, which are named by \citet{aubin2021linear} as Example-2/2s/3/3s. Example-2 and 3 are two binary classification tasks of Gaussian linear data generated in processes similar to the setup of Sec. \ref{sec:setup}, and they have identity transformation, $R=I$ (see the definition of $R$ in \eqref{eq:def-linear-transform}), while Example 2s/3s are their counterparts with $R$ as a random transformation matrix. However, Example-3/3s do not satisfy Assumption \ref{assum:non-degenerate-cov}, thus cannot properly examine our ISR-Cov. Hence, we construct variants of Example-3/3s satisfying Assumption \ref{assum:non-degenerate-cov}, which we name as Example-3'/3s', respectively. We provide specific details of these benchmarks in \cref{supp:exp:synthetic-setup}.

\textit{Example-2}: The data generation process for Example-2 follows the Structual Causal Model \citep{peters2015causal}, where $P(Y|\mu_c)$ is invariant across environments.

\textit{Example-3}: It is similar to our Gaussian setup in \ref{sec:setup}, except that $\sigma_e\equiv \sigma_c=0.1$, breaking Assumption \ref{assum:non-degenerate-cov}. In this example, $P(\mu_c|Y)$ is invariant across environments

\textit{Example-3'}: We modify Example-3 slightly such that $\sigma_c=0.1$ and $\sigma_e \sim \mathrm{Unif}(0.1, 0.3)$. All the rest settings are identical to Example-3.

\textit{Example-2s/3s/3s'}: A random orthonormal projection matrix $R = [A,~B] \in \bR^{d\times d}$ (see the definition in \eqref{eq:def-linear-transform}) is applied to the original Example-2/3/3' to scramble the invariant and spurious latent feature, leading to Example-2s/3s/3s' with observed data in the form of $x = Az_c + B z_e$.

\paragraph{Implementation}
For baseline algorithms, we directly adopt their implementations by \citet{aubin2021linear}. We implement ISRs following \cref{algo:isr-mean} and \ref{algo:isr-cov}, where the last step of fitting predictors is done by the ERM implementation of \citet{aubin2021linear}, which optimizes the logistic loss with an Adam optimizer \citep{adam}. More details are provided in \cref{supp:exp}.

\paragraph{Evaluation Procedures}~
Following \citet{aubin2021linear}, we fix $d\mathrm{=}10$, $d_c \mathrm{=} 5$, $d_s\mathrm{=}5$, and increase $E$, the number of training environments, from 2 to 10, with 10K observed samples per environment. Each algorithm trains a linear predictor on these training data, and the predictor is evaluated in $E$ test environments, each with 10K data. The test environments are generated analogously to the training ones, while the spurious features $z_e$ are randomly shuffled across examples within each environment. The mean classification error of the trained predictor over $E$ test environments is evaluated.

\paragraph{Empirical Comparisons}~
We compare our ISRs with several algorithms implemented in \citet{aubin2021linear}, including IRMv1, IGA (an IRM variant by \citet{koyama2020out}), ERM and Oracle (the optimal invariant predictor) on the datasets. We repeat the experiments over 50 random seeds and plot the mean errors of the algorithms. Fig. \ref{fig:linear-unit-tests} shows the results of our experiment on these benchmarks: 
\textbf{a)} On Example-2/2s, our ISRs reach the oracle performance with a small $E$ (number of training environments), significantly outperforming other algorithms.
\textbf{b)} On Example-3/3s, ISRs reach the oracle performance as $E > 5=d_s$, while IRM or others need more environments to match the oracle. 
\textbf{c)} On Example-3'/3s', ISR-Cov matches the oracle as $E \geq 2$, while the performance of all other algorithms does not differ much from that of Example-3/3s.

\paragraph{Conclusions}~ Observing these results, we can conclude that:
\textbf{a)} ISR-Mean can stably match the oracle as $E > d_s$, validating the environment complexity proved in \cref{thm:isr-mean}. 
\textbf{b)} ISR-Cov matches the oracle as $E\geq 2$ in datasets satisfying Assumption \ref{assum:non-degenerate-cov} (i.e., Example-2/2s/3'/3s'), corroborating its environment complexity proved in \cref{thm:isr-cov}.
% In datasets that break Assumption \ref{assum:non-degenerate-cov}, ISR-Cov can still match ISR-Mean, showing a linear environment complexity.

\begin{figure*}[ht!]
\begin{center}
\centerline{\includegraphics[width=1\linewidth]{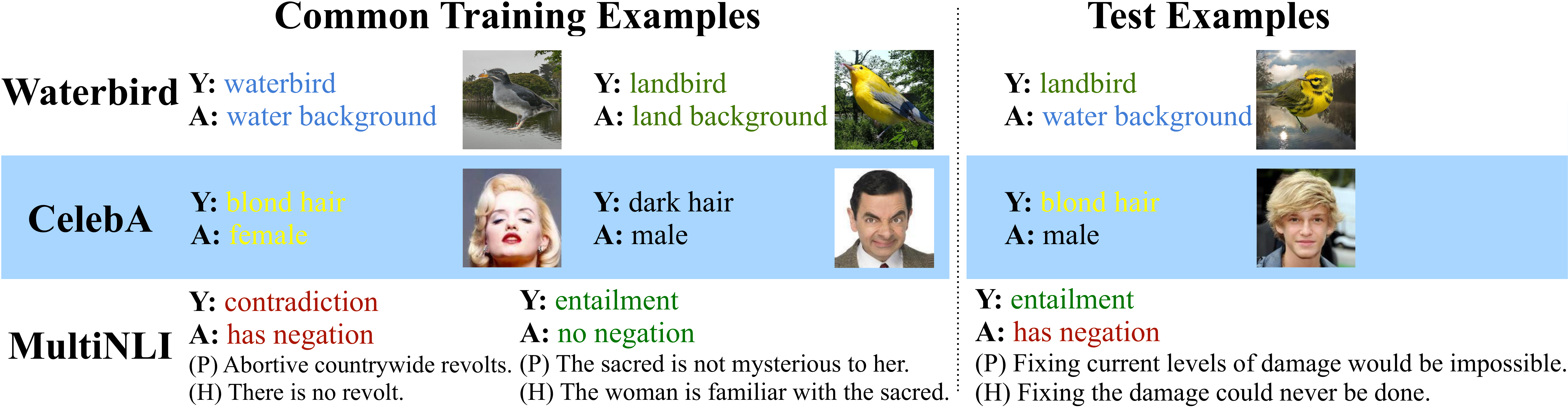}}
\vskip -0.1in
\caption{Representative examples of the three real datasets we use. The spurious correlation between the label (\textbf{Y}) and the attribute (\textbf{A}) in the training data does not hold in the test data.
}
\label{fig:real-datasets}
\end{center}
\vspace{-1em}
\end{figure*}

\vspace{.5em}
\subsection{Real Datasets} 

\begin{table*}[ht!]
\centering
    \resizebox{.99\textwidth}{!}{%
    \centering  
    \setlength{\tabcolsep}{4.5pt}
\begin{tabular}{@{}cclccccccc@{}}
\toprule
\multicolumn{1}{c}{\textbf{Dataset}}    & \multicolumn{1}{c}{\textbf{Backbone}}&
\multicolumn{1}{c}{\textbf{Algorithm}} &
\multicolumn{3}{c}{\textbf{Average Accuracy}} & 
\multicolumn{3}{c}{\textbf{Worst-Group Accuracy}} \\ \cmidrule(l){4-6} \cmidrule(l){7-9} 
& \multicolumn{2}{c}{\text{~}}&\textbf{ Original}& \textbf{ ISR-Mean}& \textbf{ ISR-Cov}& \textbf{ Original}& \textbf{ ISR-Mean} &\textbf{ ISR-Cov}\\ 
\midrule
\multicolumn{1}{c}{Waterbirds} 
& ResNet-50 &ERM& 86.66$\pm$0.67& 87.87$\pm$0.80& \textbf{90.47$\pm$0.33} ~&~
62.93$\pm$5.37& 76.10$\pm$1.11& \textbf{82.46$\pm$0.55}\\
& & Reweighting & 91.49$\pm$0.46& \textbf{91.77$\pm$0.52}& 91.63$\pm$0.44 ~&~
87.69$\pm$0.53& 88.02$\pm$0.42& \textbf{88.67$\pm$0.55}\\
& & GroupDRO& 92.01$\pm$0.33& 91.74$\pm$0.35& \textbf{92.25$\pm$0.27} ~&~
90.79$\pm$0.47& 90.42$\pm$0.61& \textbf{91.00$\pm$0.45}\\ 
\midrule
\multicolumn{1}{c}{CelebA}
&ResNet-50& ERM& \textbf{95.12$\pm$0.34}& 94.34$\pm$0.12 &90.12$\pm$2.59 ~&~
46.39$\pm$2.42& 55.39$\pm$6.13& \textbf{79.73$\pm$5.00}\\
&& Reweighting & \textbf{91.45$\pm$0.50}& 91.38$\pm$0.51 & 91.24$\pm$0.35 ~&~
84.44$\pm$1.66& \textbf{90.08$\pm$0.50}& 88.84$\pm$0.57\\
&& GroupDRO&\textbf{91.82$\pm$0.27}& 91.82$\pm$0.27& 91.20$\pm$0.23 ~&~
88.22$\pm$1.67& \textbf{90.95$\pm$0.32}& 90.38$\pm$0.42\\ 
\midrule
\multicolumn{1}{c}{MultiNLI}
&BERT& ERM&\textbf{82.48$\pm$0.40}& 82.11$\pm$0.18& 81.28$\pm$0.52 ~&~
65.95$\pm$1.65& 72.60$\pm$1.09& \textbf{74.21$\pm$2.55} \\
&& Reweighting & \textbf{80.82$\pm$0.79} & 80.53$\pm$0.88& 80.73$\pm$0.90 ~&~ 
64.73$\pm$0.32& \textbf{67.87$\pm$0.21}& 66.34$\pm$2.46\\
&& GroupDRO&\textbf{81.30$\pm$0.23}& 81.21$\pm$0.24& 81.20$\pm$0.24 ~&~
78.43$\pm$0.87& \textbf{78.95$\pm$0.95}& 78.91$\pm$0.75\\ \bottomrule
\end{tabular}}
% \vspace{-1em}
\caption{Test accuracy(\%) with standard deviation of ERM, Re-weighting and GroupDRO over three datasets. We compare the accuracy of original trained classifiers vs. ISR-Mean post-processed classifiers. The average accuracy and the worst-group accuracy are both presented. Bold values mark the higher accuracy over Original vs. ISR-Mean for a given algorithm (e.g., ERM) and a specific metric (e.g., Average Acc.).} 
    \label{tab:real-datasets}

\end{table*}

 We adopt three datasets that \citet{sagawa2019distributionally} proposes to study the robustness of models against spurious correlations and group shifts. See Fig. \ref{fig:real-datasets} for a demo of these datasets. Each dataset has multiple spurious attributes, and we treat each spurious attribute as a single environment.

\textit{Waterbirds} \citep{sagawa2019distributionally}: This is a image dataset built from the CUB \citep{wah2011caltech} and Places \citep{zhou2017places} datasets. The task of this dataset is the classification of waterbirds vs. landbirds. Each image is labelled with class $y\in \mathcal Y = \{\textit{waterbird, landbird}\}$ and environment $e\in \mathcal E= \{\textit{water background, land background}\}$. \citet{sagawa2019distributionally} defines 4 groups\footnote{Notice that the definition of environment in this paper is different from the definition of group in \citet{sagawa2019distributionally}.} by $\mathcal G = \mathcal Y \times \mathcal E$. There are 4795 training samples, and smallest group (waterbirds on land) only has 56.

\textit{CelebA} \cite{liu2015faceattributes}: This is a celebrity face dataset of 162K training samples. \citet{sagawa2019distributionally} considers a hair color classification task ($\mathcal Y = \{\textit{blond, dark}\}$) with binary genders as spurious attributes (i.e., $\mathcal E = \{\textit{male, female}\}$). Four groups are defined by $\mathcal G = \mathcal Y \times \mathcal E$, where the smallest group (blond-haired males) has only 1387 samples.

\textit{MultiNLI} \cite{williams2017broad}: This is a text dataset for natural language inference. Each sample includes two sentences, a hypothesis and a premise. The task is to identify if the hypothesis is contradictory to, entailed by, or neutral with the premise ($\mathcal Y = \{\textit{contradiction,neutral,entailment}\}$). \citet{gururangan2018annotation} observes a spurious correlation between $y\mathrm{=}\textit{contradiction}$ and negation words such as nobody, no, never, and nothing. Thus $\mathcal E \mathrm{=} \{\textit{no negation, negation}\}$ are spurious attributes (also environments), and 6 groups are defined by $\mathcal G \mathrm{=} \mathcal Y \mathrm{\times} \mathcal E$. There are 20K training data, while the smallest group (entailment with negations) has only 1521.

\paragraph{Implementation}~ We take three algorithms implemented by \citet{sagawa2019distributionally}: ERM, Reweighting, and GroupDRO. First, for each dataset, we train neural nets with these algorithms using the code and optimal hyper-parameters provided by \citet{sagawa2019distributionally} implementation, and early stop models at the epoch with the highest worst-group validation accuracy. Then, we use the hidden-layers of the trained models to extract features of training data, and fit ISR-Mean/Cov to the extracted features. Finally, we replace the original last linear layer with the linear classifier provided by ISR-Mean/Cov, and evaluate it in the test set. More details are provided in \cref{supp:exp}.

\textbf{Empirical Comparisons}~
We compare trained models with the original classifier vs. ISR-Mean/Cov post-processed classifiers over three datasets. Each experiment is repeated over 10 random seeds. From the results in \cref{tab:real-datasets}, we can observe that: \textbf{a)} ISRs can improve the worst-group accuracy of trained models across all dataset-algorithm choices. \textbf{b)} Meanwhile, the average accuracy of ISR-Mean/Cov is maintained around the same level as the original classifier.

\begin{table*}[t]
\centering
    \resizebox{.99\textwidth}{!}{%
    \centering  
    \setlength{\tabcolsep}{4.5pt}
\begin{tabular}{@{}cclccccccc@{}}
\toprule
\multicolumn{1}{c}{\textbf{Dataset}}    & \multicolumn{1}{c}{\textbf{Backbone}}&
\multicolumn{1}{c}{\textbf{Algorithm}} &
\multicolumn{3}{c}{\textbf{Average Accuracy}} & 
\multicolumn{3}{c}{\textbf{Worst-Group Accuracy}} \\ \cmidrule(l){4-6} \cmidrule(l){7-9} 
& \multicolumn{2}{c}{\text{~}}&\textbf{\small Linear Probing}& \textbf{\small ISR-Mean}& \textbf{\small ISR-Cov}& \textbf{\small Linear Probing}& \textbf{\small  ISR-Mean} &\textbf{\small ISR-Cov}\\ 
\midrule
\multicolumn{1}{c}{Waterbirds} 
& CLIP (ViT-B/32) &ERM& 76.42$\pm$0.00& \textbf{90.27$\pm$0.09}& 76.80$\pm$0.01 ~&~
52.96$\pm$0.00& \textbf{71.75$\pm$0.39}& 55.76$\pm$0.00\\
&&Reweighting & 87.38$\pm$0.09& \textbf{88.23$\pm$0.12}& 88.07$\pm$0.05 ~&~
82.51$\pm$0.27& \textbf{85.13$\pm$0.22}& 83.33$\pm$0.00\\
\bottomrule
\end{tabular}}
% \vspace{-1em}
\caption{Evaluation with CLIP-pretrained vision transformers. We compare ISR-Mean/ISR-Cov vs. linear probing in the Waterbird dataset, and report the test accuracy (\%) with standard deviation.} 
    \label{tab:CLIP}

\end{table*}

\vspace{.3em}
\subsubsection{Reduced Requirement of Environment Labels} 
\begin{figure}[t!]
% \begin{center}
\vspace{-.5em}
\includegraphics[width=.95\columnwidth]{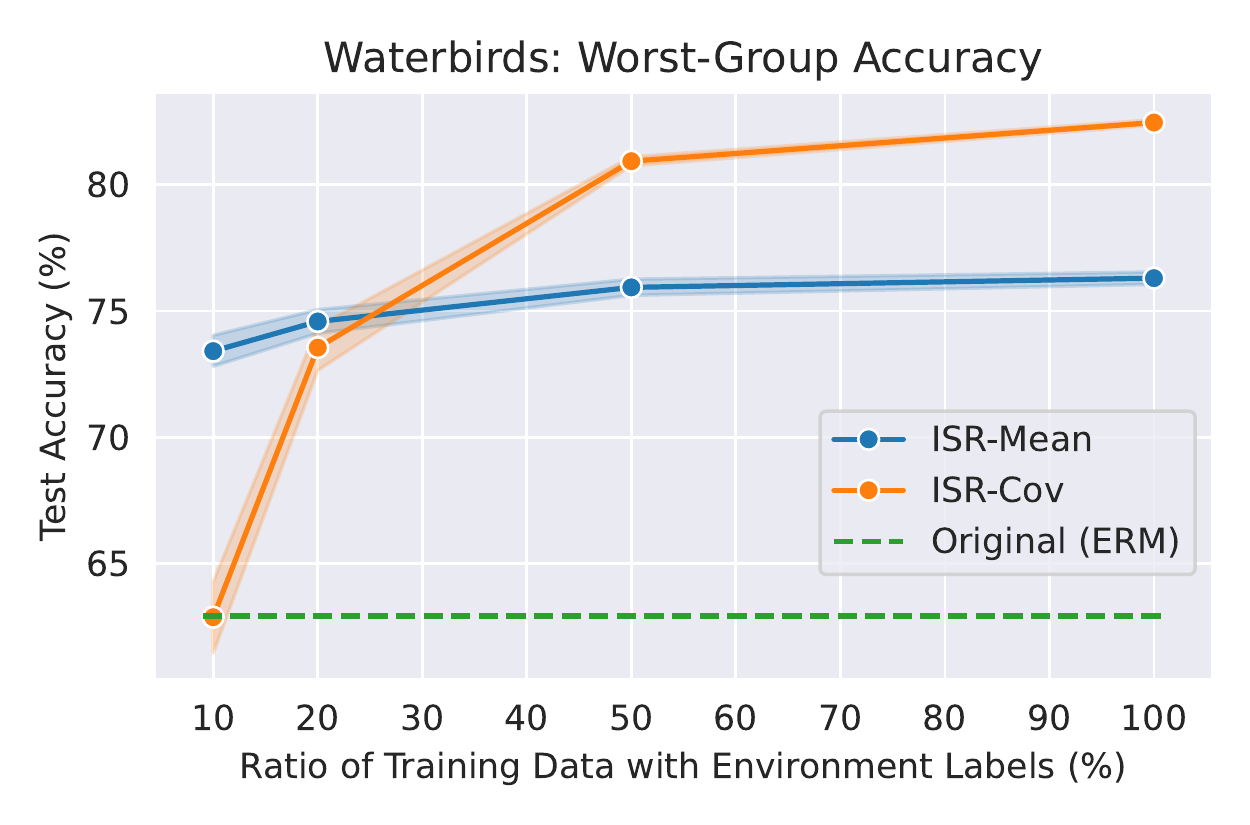}
\vspace{-1em}
\caption{Applying ISR-Mean/ISR-Cov to ERM-trained models with partially available environment labels in the Waterbirds dataset. The shading area indicates the 95\% confidence interval for mean accuracy.
}
% \vspace{-1em}
\label{fig:partial-env}
% \end{center}
% \vskip -0.2in
\end{figure}

Algorithms such as GroupDRO are successful, but they require each training sample to be presented in the form $(x,y,e)$, where the environment label $e$ is usually not available in many real-world datasets. Recent works such as \citet{liu2021just} try to relieve this requirement. To this end, we conduct another experiment on Waterbirds to show that ISRs can be used in cases where only a part of training samples are provided with environment labels. Adopting the same hyperparameter as that of Table \ref{tab:real-datasets}, we reduce the available environment labels from $100\%$ to $10\%$ (randomly sampled), and apply ISR-Mean/Cov on top of ERM-trained models with the limited environment labels. We repeat the experiment over 10 runs for each of 10 ERM-trained models, and plot the mean accuracy in Fig. \ref{fig:partial-env}. We can observe that \textbf{a)} even with only $10\%$ environment labels, the worst-group accuracy of ISR-Mean attains $73.4\%$, outperforming the original ERM-trained classifier by a large margin of $10.5\%$, and \textbf{b)} with $50\%$ environment labels, the worst-group accuracy of ISR-Cov becomes $80.9\%$, surpassing the original classifier by $18.0\%$. The compelling results demonstrate another advantage of our ISRs, the \textit{efficient utilization of environment labels}, which indicates that ISRs can be useful to many real-world datasets with only partial environment labels. \looseness=-1

\subsubsection{Applying ISRs to Pretrained Feature Extractors}
It is recently observed that CLIP-pretrained models \citep{CLIP} have impressive OOD generalization ability across various scenarios \citep{miller2021accuracy,wortsman2022robust,kumar2022finetuning}. Also, \citet{kumar2022finetuning} shows that over a wide range of OOD benchmarks, linear probing (i.e., re-training the last linear layer only) could obtain better OOD generalization performance than finetuning all parameters for CLIP-pretrained models. Notice that ISR-Mean \& ISR-Cov also re-train last linear layers on top of provided feature extractors, thus our ISRs can be used as substitutes for linear probing on CLIP-pretrained models. We empirically compare ISR-Mean/Cov vs. linear probing for a CLIP-pretrained vision transformer (ViT-B/32) in the Waterbirds dataset. As Table \ref{tab:CLIP} shows, ISRs outperform linear probing in terms of both average and worst-group accuracy, and the improvement that ISR-Mean obtains is more significant than that of ISR-Cov. This experiment indicates that our ISRs could be useful post-processing tools for deep learning practitioners who frequently use modern pre-trained (foundation) models \citep{foundation-models}.

\section{Conclusion}\label{sec:conclusion}
% \vspace{-.2em}
In this paper, under a common data generative model in the literature, we propose two algorithms, ISR-Mean and ISR-Cov, to achieve domain generalization by recovering the invariant-feature subspace. We prove that ISR-Mean admits an $\cO(d_s)$ environment complexity and ISR-Cov obtains an $\cO(1)$ environment complexity, the minimum environment complexity that any algorithm can hope for. Furthermore, both algorithms are computationally efficient, free of local minima, and can be used off-the-shelf as a post-processing method over features learned from existing models. Empirically, we test our algorithms on synthetic benchmarks and demonstrate their superior performance when compared with other domain generalization algorithms. We also show that our proposed algorithms can be used as post-processing methods to increase the worst-case accuracy of (pre-)trained models by testing them on three real-world image and text datasets.

\section*{Acknowledgements}
This work is partially supported by NSF grant No.1910100, NSF CNS No.2046726, C3 AI, and the Alfred P. Sloan Foundation. BL and HZ would like to thank the support from a Facebook research award.

\newpage
\bibliography{reference.bib}
\bibliographystyle{icml2022}

%%%%%%%%%%%%%%%%%%%%%%%%%%%%%%%%%%%%%%%%%%%%%%%%%%%%%%%%%%%%%%%%%%%%%%%%%%%%%%%
%%%%%%%%%%%%%%%%%%%%%%%%%%%%%%%%%%%%%%%%%%%%%%%%%%%%%%%%%%%%%%%%%%%%%%%%%%%%%%%
% APPENDIX
%%%%%%%%%%%%%%%%%%%%%%%%%%%%%%%%%%%%%%%%%%%%%%%%%%%%%%%%%%%%%%%%%%%%%%%%%%%%%%%
%%%%%%%%%%%%%%%%%%%%%%%%%%%%%%%%%%%%%%%%%%%%%%%%%%%%%%%%%%%%%%%%%%%%%%%%%%%%%%%
\newpage
\appendix
\onecolumn

\section{Proof}\label{supp:proof}

\subsection{Proof of \cref{thm:isr-mean}}
\begin{proof}
From \eqref{eq:def-M}, we know
\begin{align}
    \mathcal M \mathrm{\coloneqq} \begin{bmatrix}
    \bar{x}_1^\T\\
    \vdots\\
    \bar{x}_E^\T
    \end{bmatrix} \mathrm{=} \begin{bmatrix}
    \mu_c^\T A^\T \mathrm{+} \mu_1^\T B^\T\\
    \vdots\\
    \mu_c^\T A^\T \mathrm{+} \mu_E^\T B^\T\end{bmatrix}
    \mathrm{=}{\overbrace{\begin{bmatrix}
    \mu_c^\T ~~\mu_1^\T\\
    \vdots~~~~~\vdots\\
    \mu_c^\T ~~ \mu_E^\T
    \end{bmatrix}}^{\mathcal U^\T \coloneqq}} R^\T = (R\mathcal U )^\T
\end{align}
where $\mathcal U \coloneqq \begin{bmatrix}
\mu_c & \dots & \mu_c \\
\mu_1 & \dots & \mu_E 
\end{bmatrix}\in \bR^{d \times E}
$

If $E\leq d_s$, Assumption \ref{assum:non-degenerate-mean} guarantees that $\{\mu_1,\dots, \mu_E\}$ are linearly independent almost surely. Then, we have $\rank(\mathcal U) = E$. As $E > d_s$, since the first $d_c$ rows of $\mathcal U$ are the same, the rank of $\mathcal U$ is capped, i.e., $\rank(\mathcal U) = d - d_c = d_s$.

The mean-subtraction step of PCA compute the sample-mean
\begin{align}
\wt x = \frac{1}{E}\sum_{e=1}^E\bar x_e = A \mu_c + B \left(\frac{1}{E}\sum_{e=1}^E \mu_e\right) = A \mu_c + B \bar \mu,
\end{align}
where $\bar \mu \coloneqq \frac{1}{E}\sum_{e=1}^E \mu_e $, and then subtracts $\wt x^\T$ off each row of $\mathcal M$ to obtain
\begin{align}
    \wt {\mathcal M} \mathrm{\coloneqq} \begin{bmatrix}
    \bar{x}_1^\T - \wt x^\T \\
    \vdots\\
    \bar{x}_E^\T - \wt x^\T
    \end{bmatrix} \mathrm{=} \begin{bmatrix}
    (\mu_1-\bar \mu)^\T B^\T \\
    \vdots\\
    (\mu_E-\bar \mu)^\T B^\T\end{bmatrix}
    \mathrm{=}{\overbrace{\begin{bmatrix}
    \mu_1^\T - \bar \mu^\T\\
    \vdots\\
    \mu_E^\T - \bar \mu^\T
    \end{bmatrix}}^{\wt{\mathcal U}^\T \coloneqq}} B^\T = (B \wt {\mathcal U} )^\T\in \bR^{ d_s \times d}
\end{align}
where $\wt{\mathcal U} \coloneqq \begin{bmatrix}
\mu_1 - \bar \mu & \dots & \mu_E - \bar \mu 
\end{bmatrix}\in \bR^{d_s \times E}
$

Similar to the analysis of $\mathcal U$ above, we can also analyze the rank of $\wt{\mathcal U}$ in the same way. However, different from $\cU$, we have $\rank(\wt\cU) = \min\{d_s, E-1\}$, where the $-1$ comes from the constraint $\sum_{e=1}^E (\mu_e - \bar \mu) = 0$ that is put by the mean-subtraction.

Suppose $E\geq d_s +1$, then $\rank(\wt\cU) = d_s$. The next step of PCA is to eigen-decompose the sample covariance matrix
\begin{align*}
    \frac{1}{E}\wt\cM^\T \wt\cM =  \frac{1}{E} (B \wt {\mathcal U} ) (B \wt {\mathcal U} )^\T &= \frac{1}{E} B (\wt {\mathcal U} \wt {\mathcal U}^\T ) B^\T\\
    &= \frac{1}{E} 
    \begin{bmatrix}
    A&B
    \end{bmatrix}
    \begin{bmatrix}
    \textbf{0}_{d_c \times d_c} & \textbf{0}_{d_c \times d_s}\\
    \textbf{0}_{d_s\times d_c} & \wt {\mathcal U} \wt {\mathcal U}^\T
    \end{bmatrix}
    \begin{bmatrix}
    A^\T\\
    B^\T
    \end{bmatrix}\in \bR^{d \times d}\eq \label{eq:supp:proof:isr-mean:cov-eigendecom}
\end{align*}
where $\textbf{0}_{n\times m}$ is a $n\times m$ matrix with all zero entries, and $\wt {\mathcal U} \wt {\mathcal U}^\T \in \bR^{d_s \times d_s}$ is full-rank because  $\rank(\wt\cU) = d_s$. 

Combining with the fact that $R=[A B]$ is full-rank (ensured by Assumption \ref{assum:full-rank-transform}), we know that $\rank(\frac{1}{E}\wt\cM^\T \wt\cM) = d_s$. Therefore, $\frac{1}{E}\wt\cM^\T \wt\cM$ is positive-definite.

As a result, the eigen-decomposition on $\frac{1}{E}\wt\cM^\T \wt\cM$ leads to an eigen-spectrum of $d_s$ positive values and $d_c = d - d_s$ zero eigenvalues. 

Consider ascendingly ordered eigenvalues $\{\lambda_1,\dots, \lambda_d\}$, and compose a diagonal matrix $S$ with these eigenvalues as in ascending order, i.e., $S\coloneqq\mathrm{diag}(\{\lambda_1,\dots, \lambda_d\})$. Denote the eigenvectors corresponding with these eigenvalues as $\{P_1,\dots, P_{d_c}\}$, and stack their transposed matrices as
\begin{align}
    P \coloneqq
    \begin{bmatrix}
    P_1^\T \\
    \vdots \\
    P_d^\T
    \end{bmatrix}
    \in \bR^{d\times d}
\end{align}
Then, we have the equality
\begin{align}
    \frac{1}{E} B (\wt {\mathcal U} \wt {\mathcal U}^\T ) B^\T = \frac{1}{E}\wt\cM^\T \wt\cM = P S P^\T 
\end{align}

Since the first $d_c$ diagonal entries of $S$ are all zeros and the rest are all non-zero, the dimensions of $P$ that correspond to non-zero diagonal entries of $S$ can provide us with the subspace spanned by the $d_s$ spurious latent feature dimensions, thus the rest dimensions of $P$ (i.e., the ones with zero eigenvalues) correspond to the subspace spanned by $d_c$ invariant latent feature dimensions, i.e.,
\begin{align}
    \mathrm{Span}(\{P_i^\T R: i\in[d],~ S_{ii} = 0\}) = \mathrm{Span}(\{\mathbf{\hat {1}},\dots, \mathbf{\hat d_s}\})
\end{align}
Since the diagonal entries of $S$ are sorted in ascending order, we can equivalently write it as
\begin{align}\label{eq:thm:inv-subspace-recovery:span-equality:supp}
    \mathrm{Span}(\{P_1^\T R, \dots, P_{d_c}^\T R\}) = \mathrm{Span}(\{\mathbf{\hat {1}},\dots, \mathbf{\hat d_c}\})
\end{align}
Then, by \cref{prop:optimal-inv-pred} (i.e., Definition 1 of \citet{risks-of-IRM}) and Lemma F.2 of \citet{risks-of-IRM}, for the ERM predictor fitted to all data that are projected to the recovered subspace, we know it is guaranteed to be the optimal invariant predictor (defined in \cref{prop:optimal-inv-pred} as Eq. \eqref{eq:optimal-inv-pred}).
\end{proof}

\subsection{Proof of \cref{thm:isr-cov}}
\begin{proof}
From \eqref{eq:delta-sigma-expression}, we know
\begin{align}
\Delta \Sigma \coloneqq \Sigma_{e_1} - \Sigma_{e_2} = (\sigma_{e_1}^2 - \sigma_{e_2}^2)BB^\T \in \bR^{d \times d}
\end{align}
Assumption \ref{assum:non-degenerate-cov} guarantees that $\sigma_{e_1}^2 - \sigma_{e_2}^2 \neq 0$, and Assumption \ref{assum:full-rank-transform} ensures that $\rank(B) = d_s$. Thus, eigen-decomposition on $\Delta \Sigma$ leads to exactly $d_c$ zero eigenvalues and $d_s = 1- d_c$ non-zero eigenvalues. One just need to follow the same steps as \eqref{eq:supp:proof:isr-mean:cov-eigendecom}-\eqref{eq:thm:inv-subspace-recovery:span-equality:supp} to finish the proof.
\end{proof}
%%%%%%%%%%%%%%%%%%%%%%%%%%%%%%%%%%%%%%%%%%%%%%%%%%%%%%%%%%%%%%%%%%%%%%%%%%%%%%%
%%%%%%%%%%%%%%%%%%%%%%%%%%%%%%%%%%%%%%%%%%%%%%%%%%%%%%%%%%%%%%%%%%%%%%%%%%%%%%%

\section{Experimental Details}\label{supp:exp}

\subsection{Setups of Synthetic Datasets}\label{supp:exp:synthetic-setup}

\paragraph{Example-2}
This is a binary classification task that imitate the following example inspired by \citet{IRM, beery2018recognition}: while most cows appear in grasslands and most camels appear in desserts, with small probability such relationship can be flipped. In this example, \citet{aubin2021linear} define the animals as invariant features with mean $\pm \mu_c$ and the backgrounds as spurious features with mean $\pm \mu_e$. \citet{aubin2021linear} also scale the invariant and spurious features with $\nu_c$ and $\nu_e$ respectively. To be specific, we set $\mu_{c} = \mathbf{1}_{d_c}$ (i.e., a $d_c$-dimensional vector with all elements equal to $1$) , $\mu_{e} = \mathbf{1}_{d_e}$, $\nu_c = 0.02$ and $\nu_e = 1$. For any training environment $e \in \mathcal{E}$, \citet{aubin2021linear} construct its dataset $\mathcal D_{e}$ by generating each input-label pair $(x,y)$ in the following process:
\begin{align*}
    j_e &\sim \text{Categorical}\left(p^e s^e , (1 - p^e) s^e, p^e (1-s^e) , (1 - p^e) (1-s^e)\right)\\
        z_c &\sim
        \left\{\begin{array}{lr}
            +1\cdot \left(\mu_c+\mathcal{N}_{d_c}(0, 0.1) \right) \cdot \nu_c & \text{ if } j_e \in \{1, 2\},\\
            -1\cdot \left(\mu_c + \mathcal{N}_{d_c}(0, 0.1)\right) \cdot \nu_c & \text{ if } j_e \in \{3, 4\},\\
        \end{array}\right.\\
        z_e &\sim
        \left\{\begin{array}{lr}
            +1\cdot \left(\mu_e + \mathcal{N}_{d_s}(0, 0.1) \right) \cdot \nu_e & \text{ if } j_e \in \{1, 4\},\\
            -1\cdot \left(\mu_e + \mathcal{N}_{d_s}(0, 0.1)  \right)  \cdot \nu_e & \text{ if } j_e \in \{2, 3\},
        \end{array}\right. 
\end{align*}
\begin{align*}
        z \leftarrow \begin{bmatrix}
        z_c\\
        z_e
        \end{bmatrix}, \qquad 
        y \leftarrow
        \left\{\begin{array}{lr}
        1 & \text{ if } 1_{d_c}^\T z_c > 0,\\
        0 & \text{else}
        \end{array}\right. ,\qquad 
     x = Rz \quad 
     \text{with} \quad
     R = I_{d} , 
\end{align*}
where the background probabilities are $p^{e=0} = 0.95$, $p^{e=1} = 0.97$, $p^{e=2} = 0.99$ and the animal probabilities are $s^{e=0} = 0.3$, $s^{e=1} = 0.5$, $s^{e=2} = 0.7$. If there are more than three environments, the extra environment variables are drawn according to $p^{e} \sim \textrm{Unif}(0.9, 1)$ and $s^{e} \sim \textrm{Unif}(0.3, 0.7)$.

\paragraph{Example-3}~
This is a linear version of the spiral binary classification problem proposed by \citet{parascandolo2020learning}. In this example, \citet{aubin2021linear} assign the first $d_c$ dimensions of the features with an invariant, small-margin linear decision boundary, and the reset $d_e$ dimensions have a changing, large-margin linear decision boundary. To be specific, for all environments, the $d_c$ invariant features are sampled from a distribution with a constant mean, while the means are sampled from a Gaussian distribution for the $d_e$ spurious features. In practice set $\gamma = 0.1 \cdot \mathbf{1}_{d_c}$, $\mu_e \sim \mathcal{N}(\mathbf{0}_{d_c}, I_{d_c})$, and $\sigma_c=\sigma_e=0.1$, for all environments. For any training environment $e \in \mathcal{E}$, \citet{aubin2021linear} construct its dataset $\mathcal D_{e}$ by generating each input-label pair $(x,y)$ in the following process:
\begin{align*}
    y &\sim \text{Bernoulli}\left(\frac{1}{2}\right),\\
    z_c &\sim
    \left\{\begin{array}{lr}
        \mathcal{N}(+\gamma, \sigma_c I_{d_c}) & \text{ if } y = 0,\\
        \mathcal{N}(-\gamma, \sigma_c I_{d_c}) & \text{ if } y = 1;\\
    \end{array}\right.\\
    z_e &\sim
    \left\{\begin{array}{lr}
        \mathcal{N}(+\mu_e, \sigma_e I_{d_s}) & \text{ if } y = 0,\\
        \mathcal{N}(-\mu_e, \sigma_e I_{d_s}) & \text{ if } y = 1;\\
    \end{array}\right.
\end{align*}
\begin{align*}
        z \leftarrow \begin{bmatrix}
        z_c\\
        z_e
        \end{bmatrix}, \qquad 
     x = Rz \quad 
     \text{with} \quad
     R = I_{d} , 
\end{align*}

\paragraph{Example-3'} As explained in \cref{sec:lut}, in order to make Example-3 follow Assumption \ref{assum:non-degenerate-cov}, we slightly modify the variance of the  features in Example-3 so that $\sigma_c=0.1$ and $\sigma_e\sim \mathrm{Unif}(0.1, 0.3)$. All the rest settings are unchanged.

\paragraph{Example-2s/3s/3s'} In order to increase the difficulty of the tasks, we defined the ``scrambled`` variations of the three problems described above. To build the scrambled variations, we no longer use the identity matrix $I_d$ as the transformation matrix $R$; instead, a random orthonormal matrix $R\in\mathbb{R}^{d\times d}$ is applied to the features for all environments $e\in\mathcal{E}$. The random transformation matrix is built from a Gaussian matrix (see the code \url{https://github.com/facebookresearch/InvarianceUnitTests} of \citet{aubin2021linear} for details).

\subsection{Experiments on Synthetic Datasets}

\paragraph{Code} We adopt the codebase of Linear Unit-Tests \citep{aubin2021linear}, which provide implementations of Example-2/2s/3/3s and multiple algorithms (including IRMv1, IGA, ERM, Oracle). This codebase is released at \url{https://github.com/facebookresearch/InvarianceUnitTests}.

\paragraph{Hyper-parameters} Similar to \citet{aubin2021linear}, we perform a hyper-parameter search of 20 trials. For each trial, we train the algorithms on the training split of all environments for 10K full-batch Adam \citep{adam} iterations. We run the search for ISR-Mean and ISR-Cov algorithms on all examples, and run the search for ERM, IGA \citep{koyama2020out}, IRMv1 \citep{IRM} and Oracle on Example-3' and Example-3s'. We choose the hyper-parameters that minimize the mean error over the validation split of all environments. The experiment results for ERM, IGA, IRMv1 and Oracle on Example-2, Example-2s, Example-3 and Example-3s are from \citep{aubin2021linear}, thus we do not perform any search on them.

\subsection{Experiments on Real Datasets}
\paragraph{Training} We directly use models, hyper-parameters and running scripts provided by authors of \citet{sagawa2019distributionally} in \url{https://github.com/kohpangwei/group_DRO}. Specifically, they use ResNets \citep{resnet} for Waterbirds and CelebA, and deploy BERT \citep{bert} for MultiNLI. We train the neural nets following the official running scripts provided in \url{https://worksheets.codalab.org/worksheets/0x621811fe446b49bb818293bae2ef88c0} over 10 random seeds for Waterbirds/CelebA/MultiNLI. Each run leads to one trained neural network selected on the epoch with the highest worst-group validation accuracy.

\paragraph{ISR-Mean} There are only $E=2$ environments for Waterbirds, CelebA and MultiNLI and ISR-Mean can only identify a $\min\{E-1, d_s\}$-dimensional spurious subspace. Thus we assume $d_s=1$ for the three datasets when applying ISR-Mean. 

\paragraph{ISR-Cov} For real datasets, we do not know the $d_s$ of the learned features, thus we have to treat $d_s$ as a hyperparameter for Algorithm \ref{algo:isr-cov}. 

\paragraph{Numerical Techniques}
The feature space of learned models is usually of a high dimension (e.g., 2048 for ResNet-50 in Waterbirds/CelebA), while the features of training data usually live in a subspace (approximately). Thus, we typically apply dimension reduction to features through a PCA. Then, to overcome some numerical instability challenges, we modify Algorithm \ref{algo:isr-mean} \& \ref{algo:isr-cov} slightly: Instead of directly identifying the invariant-feature subspace as Algorithm \ref{algo:isr-mean} \& \ref{algo:isr-cov} suggest, we apply ISR-Mean/Cov in an equivalent approach: we first identify the spurious-feature subspace, and then reduces scales of features along the spurious-feature subspace. The final step of fitting linear predictors in Algorithm \ref{algo:isr-mean}/\ref{algo:isr-cov} is done by logistic regression solver provided in scikit-learn \citet{sklearn}. But in some cases, we find that directly adapting the original predictor of the trained model also yields good performance. See more details in \url{https://github.com/Haoxiang-Wang/ISR}.

\end{document}